%% file: main.tex
\documentclass[letterpaper, 10 pt, conference]{ieeeconf}  

\IEEEoverridecommandlockouts                              

\usepackage [misc]{ifsym}

\usepackage{times}
\usepackage{epsfig}
\usepackage{graphicx}
\usepackage{amsmath}
\usepackage{amssymb}

\usepackage{booktabs}
\usepackage{color}
\usepackage{xcolor}
\usepackage{tikz}
\usepackage{siunitx}
\usepackage{subcaption}

\usepackage[flushleft]{threeparttable}
\usepackage{diagbox}
\usepackage{multirow}
\usepackage{tabulary}
\usepackage{hyperref}

\input{marcos}

\title{\LARGE \bf
RaSim: A Range-aware High-fidelity RGB-D Data Simulation Pipeline \\ for Real-world Applications
}

\author{Xingyu Liu$^{*}$,
Chenyangguang Zhang$^{*}$,
Gu Wang,
Ruida Zhang,
and Xiangyang Ji$^{**}$  
\thanks{
Xingyu Liu, Chenyangguang Zhang, Ruida Zhang, and Xiangyang Ji are with the Department of Automation, Tsinghua University, Beijing, 100084, China, and also with BNRist, Beijing, 100084, China.
E-mail: \{liuxy21,zcyg22,zhangrd23\}@mails.tsinghua.edu.cn, xyji@tsinghua.edu.cn.
}
\thanks{Gu Wang is with the Lab for High Technology, Tsinghua University, Beijing, 100084, China.
E-mail: guwang12@gmail.com.}
\thanks{$^{*}$: Xingyu Liu and Chenyangguang Zhang have equally contributed.}
\thanks{$^{**}$: Corresponding author.}
}

\begin{document}

\maketitle
\thispagestyle{empty}
\pagestyle{empty}

\input{contents/0_abstract}
\input{contents/1_intro}
\input{contents/2_related_work}
\input{contents/3_pipeline}
\input{contents/4_network}
\input{contents/5_exp}
\input{contents/6_conclusion}
\input{contents/7_acknowledgment}

\bibliographystyle{IEEEtran}
\bibliography{main}

\end{document}

%% file: marcos.tex
\newcolumntype{L}[1]{>{\raggedright\let\newline\\\arraybackslash\hspace{0pt}}m{#1}}
\newcolumntype{C}[1]{>{\centering\let\newline\\\arraybackslash\hspace{0pt}}m{#1}}
\newcolumntype{R}[1]{>{\raggedleft\let\newline\\\arraybackslash\hspace{0pt}}m{#1}}

\newcommand{\app}{\raise.17ex\hbox{$\scriptstyle\sim$}}

\newlength\savewidth\newcommand\shline{\noalign{\global\savewidth\arrayrulewidth
  \global\arrayrulewidth 1pt}\hline\noalign{\global\arrayrulewidth\savewidth}}
\newcommand{\tablestyle}[2]{\setlength{\tabcolsep}{#1}\renewcommand{\arraystretch}{#2}\centering\footnotesize}

\newcommand{\tabincell}[2]{\begin{tabular}{@{}#1@{}}#2\end{tabular}}

\newcommand{\fig}[1]{Fig.~\ref{#1}}
\newcommand{\sect}[1]{Sec.~\ref{#1}}
\newcommand{\tbl}[1]{Table~\ref{#1}}

\definecolor{darkgreen}{RGB}{114,210,115}
\definecolor{lime}{RGB}{34,139,34}
\definecolor{lxycolor}{RGB}{50,200,0}
\definecolor{rudycolor}{RGB}{102,8,116}

\makeatletter
\DeclareRobustCommand\onedot{\futurelet\@let@token\@onedot}
\def\@onedot{\ifx\@let@token.\else.\null\fi\xspace}

\makeatother

\newcommand{\etal}[0]{\textit{et al.}}
\newcommand{\ie}[0]{\textit{i.e.,}}

\newcommand{\wrt}[0]{\textit{w.r.t.~}}
\newcommand{\vs}[0]{\textit{v.s.~}}


%% file: contents/0_abstract.tex
\begin{abstract}
In robotic vision, a de-facto paradigm is to learn in simulated environments and then transfer to real-world applications, which poses an essential challenge in bridging the sim-to-real domain gap.
While mainstream works tackle this problem in the RGB domain, we focus on depth data synthesis and develop a 
\textit{R}ange-\textit{a}ware RGB-D data \textit{Sim}ulation pipeline (\textit{RaSim}).
In particular, high-fidelity depth data is generated by imitating the imaging principle of real-world sensors.
A range-aware rendering strategy is further introduced to enrich data diversity.
Extensive experiments show that models trained with RaSim can be directly applied to real-world scenarios without any finetuning and excel at downstream RGB-D perception tasks.
Data and code are available at \href{https://github.com/shanice-l/RaSim}{https://github.com/shanice-l/RaSim}.
\end{abstract}

%% file: contents/1_intro.tex
\section{INTRODUCTION}


With the advent of deep learning, neural networks have emerged as dominance for numerous 3D vision tasks, including 3D semantic segmentation~\cite{liu2022cmx,chen2021spatial,chen2020bi,sodano2023robust}, object pose estimation~\cite{posecnn,labbe2020cosypose,Wang_2021_GDRN}, and depth completion~\cite{sajjan2020clear,zhu2021rgb,hu2021penet}.
However, Convolutional Neural Networks (CNNs) and Transformers are extremely data-driven, requiring vast amounts of high-fidelity RGB-D data during the training process.
Moreover, obtaining large-scale 3D datasets and annotating their precise labels are extremely time-consuming and labor-intensive.

As a result, numerous approaches have been proposed to address the lack of real RGB-D data and labels.
One of the most effective strategies is to simulate large-scale synthetic training data using tools like Blender~\cite{blender} or OpenGL.
Oftentimes, domain randomization is also employed to ensure the diversity of the data~\cite{hagelskjaer2021bridging,kollar2022simnet}.
However, rendered images still exhibit the drawbacks of low quality and lack of physical plausibility.
Therefore, recent works have shifted their focus towards employing physically-based rendering techniques~\cite{greff2021kubric,hodan2019photorealistic,tremblay2018deep} to enhance image quality.
While substantial efforts have been invested in enhancing the fidelity of synthetic RGB data,
the sim-to-real domain gap \wrt the depth modality is still obvious.
This is because synthetic depth data is typically flawless, whereas real-world depth data is incomplete, along with blur and artifacts.

\begin{figure}[tb!]
\begin{center}  \includegraphics[width=\linewidth]{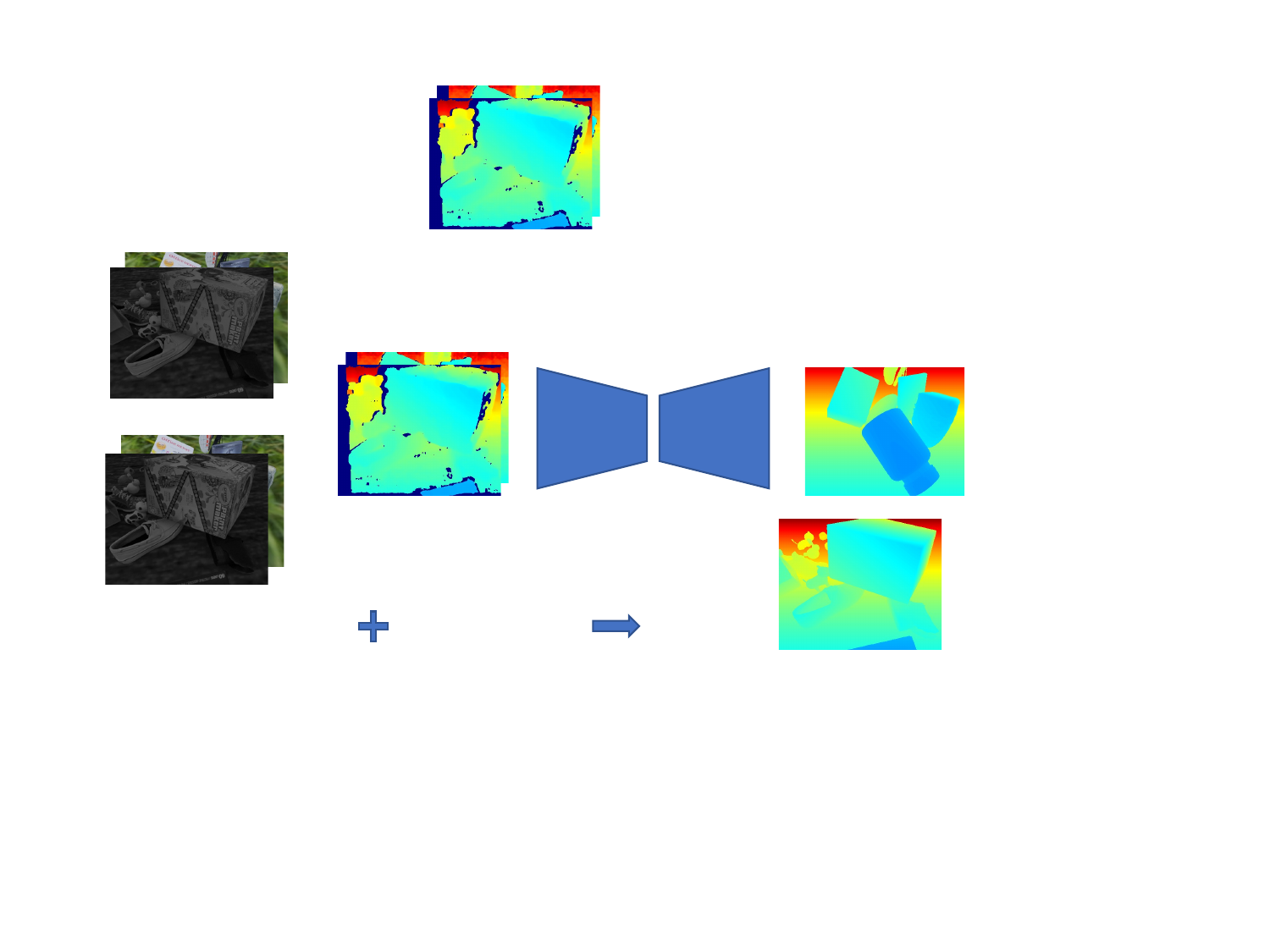}
\end{center}
  \caption{Illustration of the core idea. 
  We first generate high-fidelity simulated depth maps by imitating the imaging principle of the stereo camera, and further design a range-aware rendering strategy that renders binocular IR or RGB images according to distance to enrich data diversity.
  Then an SDRNet is devised to restore the ground-truth depth from simulated depth. 
  }
\vspace{-3mm}
\label{fig:teaser}
\end{figure}

To alleviate this problem, we introduce a \textbf{R}ange-\textbf{a}ware RGB-D data \textbf{Sim}ulation pipeline named \textbf{RaSim} to produce high-fidelity simulated 3D data.
As shown in \fig{fig:teaser}, our simulation system is grounded on imitating the imaging principle of the stereo camera based on the RealSense D400 series, as they have broad applications in both industrial and academic scenarios.
Implemented with Kubric~\cite{greff2021kubric}, we first generate large corpora of virtual scenes with photo-realistic object models, diverse backgrounds, global illuminations, and physical simulations.
Then simulated depth maps are obtained by performing the semi-global stereo-matching algorithm using binocular images.
We further devise a range-aware rendering strategy to enrich data diversity.
Specifically, the type of matching images for depth simulation varies between IR and RGB depending on the distance between the scene and the camera.
This strategy allows us to simulate nearby and distant scenes, enabling the pipeline to adapt to a wider range of application scenarios.

Supported by RaSim which randomizes over lighting and textures, we create a large-scale domain-randomized dataset that includes simulated and ground-truth depth maps, pixel-level semantic annotations, and millions of instances labeled with poses, categories, and 3D object coordinates.
To verify whether RaSim can assist in real-world applications, we train networks with the proposed RaSim dataset for two RGB-D-based perception tasks: depth completion and depth pre-training.
Firstly, a \textbf{S}imulated \textbf{D}epth \textbf{R}estoration \textbf{Net}work (\textbf{SDRNet}) is trained to repair the incomplete and noisy simulated depth map by decoding hierarchical RGB and depth features extracted with Swin Transformer~\cite{liu2021swin}.
Subsequently, inspired by the idea of \textit{masked language modeling} in natural language processing~\cite{devlin2018bert}, we consider depth restoration as a pre-training task for RGB-D-based Transformer. 
Specifically, weights pre-trained on RaSim are used to initialize the depth branch of the Transformer for facilitating various downstream tasks.
Note that we opt for Transformer over CNNs, in that the scarcity of data in the Transformer architecture is a more prominent concern.

To verify the effectiveness of RaSim, we conduct extensive experiments on two real-world datasets, \ie~ClearGrasp~\cite{sajjan2020clear} for depth completion and YCB-V~\cite{posecnn} for depth pre-training.
To sum up, our contributions are threefold:
\begin{itemize}
    \item By imitating the imaging principle of the stereo camera, we propose a RaSim pipeline to produce high-fidelity simulated depth and photo-realistic RGB-D images.
    A range-aware scene rendering strategy is further introduced to enrich the diversity of depth data.
    \item Supported by the RaSim pipeline, we create a large-scale synthetic RGB-D dataset that comprises more than 206K images across 9,835 diverse scenes.
    This dataset is equipped with physical simulations, comprehensive annotations, and the integration of domain randomization techniques.
    \item We conduct extensive experiments on two RGB-D-based perception tasks, \ie~depth completion and depth pre-training, to demonstrate the applicability of RaSim in real-world scenarios.
\end{itemize}

%% file: contents/2_related_work.tex
\section{RELATED WORK}

This work relates to two major strands of research: synthetic RGB-D dataset generation and learning from simulated environments. 

\subsection{Synthetic RGB-D Dataset Generation}

High-quality synthetic data generation plays a crucial role in 3D vision tasks since it is error-prone and labor-intensive to collect, calibrate, and annotate realistic RGB-D data.
There are various synthetic 3D dataset generation pipelines like BlenderProc~\cite{blender}, Omnidata~\cite{eftekhar2021omnidata}, OpenRooms~\cite{li2021openrooms}, and Kubric~\cite{greff2021kubric}.
However, the depth maps directly generated from these pipelines are too idealistic to adjust to real-world scenarios, since the depth collected from the real world could be noisy and incomplete.

More recently, Dai \etal~\cite{dai2022dreds} proposed a pipeline called DREDS to generate simulated depth by imitating the RealSense D415 camera following \cite{zhang2023close}.
However, DREDS faces limitations in terms of physics simulation, which results in semantic ambiguity. 
Additionally, the diversity of depth data is restrained by the ideal range (0.5 -- 2 meters) of their system.
Moreover, since DREDS is tailored for category-level pose estimation, the variety \wrt object categories are relatively scarce.
In contrast, our work focuses on generating a large-scale, photo-realistic RGB-D synthetic dataset featuring rich annotations, physical simulations, a diverse array of objects and scenes, and an extensive depth range.



\subsection{Learning from Simulated Environments}

In robotic vision, a widely adopted strategy involves training the network in simulated environments and subsequently transferring to real-world applications, such as robotic grasping~\cite{james2019sim}, pose estimation~\cite{Rad_2018_CVPR,deng2017hand3d,wang2019nocs}, depth completion~\cite{sajjan2020clear,dai2022dreds}, and scene understanding~\cite{song2017semantic,Roberts_2021_ICCV,szot2021habitat}.
Driven by this strategy, sim-to-real approaches like domain randomization and domain adaptation play a pivotal role in the learning process.
Specifically, domain randomization diversifies training data to adapt to various testing scenes~\cite{tobin2017domain,ren2019domain,tremblay2018training}, while domain adaptation leverages transfer learning techniques to align the simulated environment with the real world~\cite{jaritz2019xmuda,yang2021st3d,zhou2018unsupervised}.
In this work, we address the sim-to-real challenge from both perspectives.
On one hand, we introduce randomization in object categories, indoor and outdoor scenes, illuminations, and camera poses.
On the other hand, we simulate high-fidelity depth maps by imitating real-world sensors to adapt to real domains.

%% file: contents/3_pipeline.tex
\begin{figure*}[t!]
\begin{center}
  \vspace{1mm}
  \includegraphics[width=0.9\linewidth]{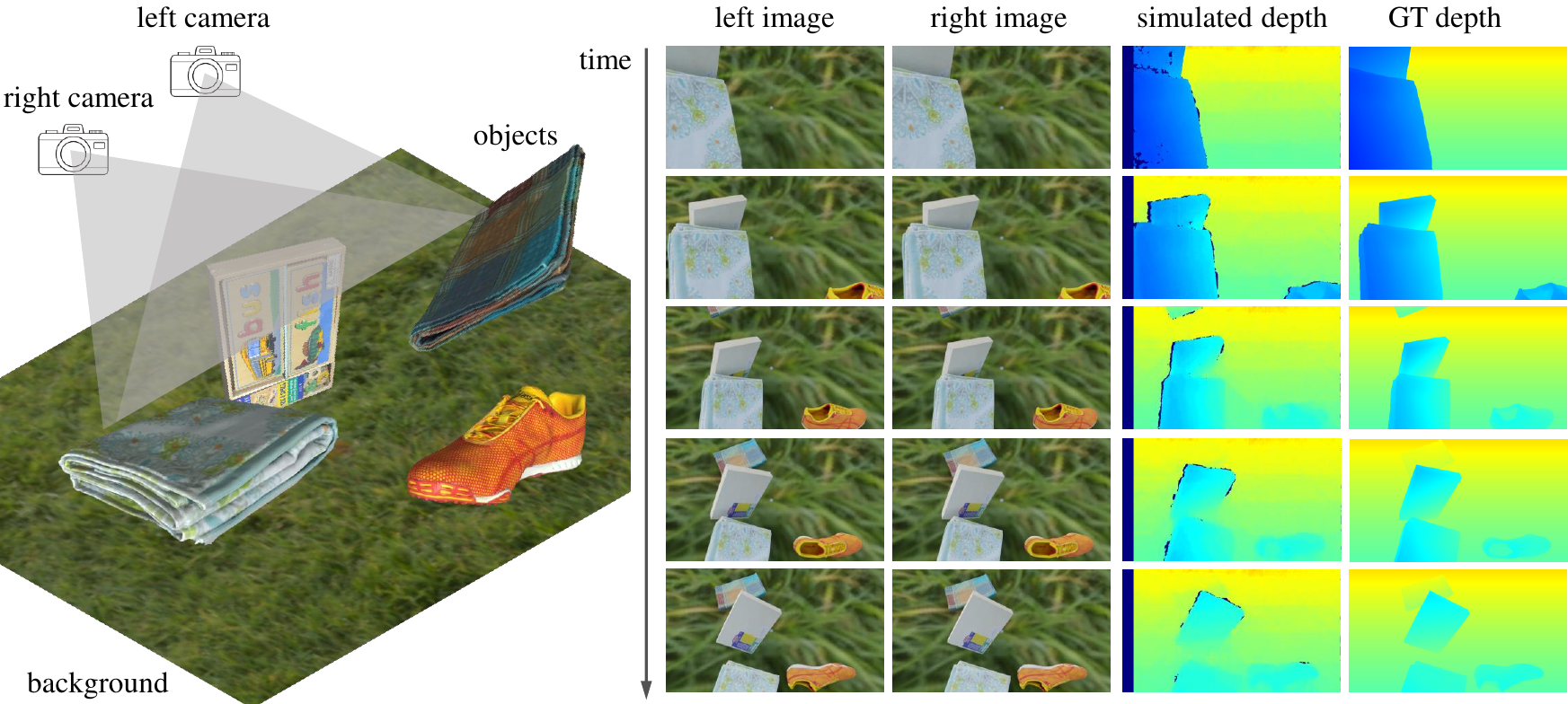}
\end{center}
  \caption{The pipeline of RaSim. 
  Given the virtual scene constructed by objects, background, and global illumination, the left and right cameras take videos under chronological physical simulation.
  Subsequently, the simulated depth maps are generated by the semi-global stereo-matching algorithm from binocular images.}
  \vspace{-3mm}
\label{fig:dataset}
\end{figure*}

\section{RANGE-AWARE RGB-D DATA SIMULATION}

\subsection{Overview}
Synthetic depth generated by the traditional pipelines is accurate, complete, and noise-free, while the depth collected from the real world is of low quality, along with blur and artifacts.
To bridge the sim-to-real domain gap, we choose to simulate active stereo depth sensors, \ie~Intel RealSense D400 Series, as they are relatively cheap and have broad applications in both industrial and academic scenarios.
RealSense D400 imaging system includes an infrared (IR) projector, stereo IR cameras with a baseline distance $\mathcal{C}_b$ and a unified focal length $\mathcal{C}_f$, as well as a central RGB camera.
After the projector emits infrared light, the stereo cameras fetch left and right IR images respectively.
Given binocular images, we could calculate the disparity value $\mathcal{D}_p$ with the semi-global matching algorithm~\cite{hirschmuller2011semi}.

Finally, the depth $z_{sim}$ is obtained as follows
\begin{equation}
z_{sim} = \frac{\mathcal{C}_b \cdot \mathcal{C}_f}{\mathcal{D}_p + \epsilon},
\end{equation}
where we set $\epsilon$ to $10^{-6}$ to avoid dividing by zero.  

\subsection{Range-aware Scene Rendering}
\label{sec:dyren}
Following \cite{dai2022dreds,zhang2023close}, we render stereo IR images by having all ambient lights emit rays in the IR spectrum with reduced intensity.
Additionally, a weak light value is added to simulate radiance from the environment.
Finally, the rendered IR images are generated in grayscale.
Despite the above pipeline generating high-fidelity depth from stereo IR images, a significant flaw is that the depth quality declines sharply when the camera is far from the scene ($\geq$2m).
It is due to the discrepancy between the left and right IR images becoming inconspicuous, along with the reduced environmental illumination, making the stereo-matching procedure error-prone.
To alleviate this problem, we propose a range-aware rendering strategy.
Recapping \fig{fig:teaser}, for nearby scenes where the camera and objects are close, we perform stereo matching with IR images. 
While for distant scenes, the matching is based on binocular RGB images with richer texture information and brighter light illumination.
Given rectified stereo images, RaSim first applies the center-symmetric census transform~\cite{zabih1994non}, followed by a semi-global stereo-matching~\cite{hirschmuller2007stereo} algorithm for disparity estimation. 
Subsequently, the disparity is further refined by median filtering and consistency checks before the conversion into depth.
This range-aware rendering strategy enriches the diversity of the dataset, yielding improvements in the versatility of the RaSim pipeline.

As shown in \fig{fig:dataset}, we denote $\mathcal{C}^{\{L, R\}}$ as the left and right stereo cameras, $\mathcal{S}$ as the virtual scene, $T$ as the total frames per rendering, $\mathbf{Z}_{gt}$ as the ground-truth depth, and $\mathbf{I}^{\{L, R\}}$ as the corresponding rendered left and right images in the format of IR or RGB. 
The rendering procedure can be formulated as follows
\begin{equation}
\begin{aligned}
\mathbf{Z}_{gt} = &  \{\mathbf{Z}_t \mid \mathbf{Z}_t = Render(\mathcal{C}^{L}, \mathcal{S}_t)\}_{t=1}^T, \\
\mathbf{I}^{\{L,R\}} = & \{\mathbf{I}_t \mid \mathbf{I}_t = Render(\mathcal{C}^{\{L,R\}}, \mathcal{S}_t)\}_{t=1}^T, \\
\mathcal{S}_t = & \mathcal{P}_t  \circ (\mathcal{O}, \mathcal{B}, \mathcal{L}).
\end{aligned}
\end{equation}
Thereby, $\mathcal{O}$ is the objects selected from the GSO~\cite{downs2022google} dataset, $\mathcal{B}$ is an indoor or outdoor background selected from either rooms textured by the CC0textures Library or scenes in Poly Haven\footnote{https://polyhaven.com/hdris},
$\mathcal{L}$ is the global illumination varying with the environment, and $\mathcal{P}_t$ stands for physical simulation working on all the assets at frame $t$.

The pipeline is implemented with Kubric~\cite{greff2021kubric}, a dataset generator interfacing with Blender~\cite{blender} and PyBullet~\cite{coumans2016pybullet}.

\subsection{RaSim Dataset}
Driven by RaSim, we create a large-scale synthetic RGB-D dataset with domain randomization and physically-based rendering techniques.
It comprises more than 206K images distributed across 9,835 diverse scenes.
Each image is annotated with pixel-level semantic information, alongside both simulated and ground-truth depth maps, and other meta information \wrt scene generation.
Moreover, one million instances featured with CAD models, poses, categories, and 3D coordinates are also included.
Thanks to rich annotations, the dataset can be applied to numerous 3D vision tasks including object manipulation~\cite{kollar2022simnet}, unseen pose estimation~\cite{labbe2023megapose,liu2022gen6d}, and 3D semantic segmentation~\cite{liu2022cmx, wang2022multimodal}.

%% file: contents/4_network.tex
\begin{figure*}[tb!]
\begin{center}
  \vspace{1mm}
  \includegraphics[width=0.9\linewidth]{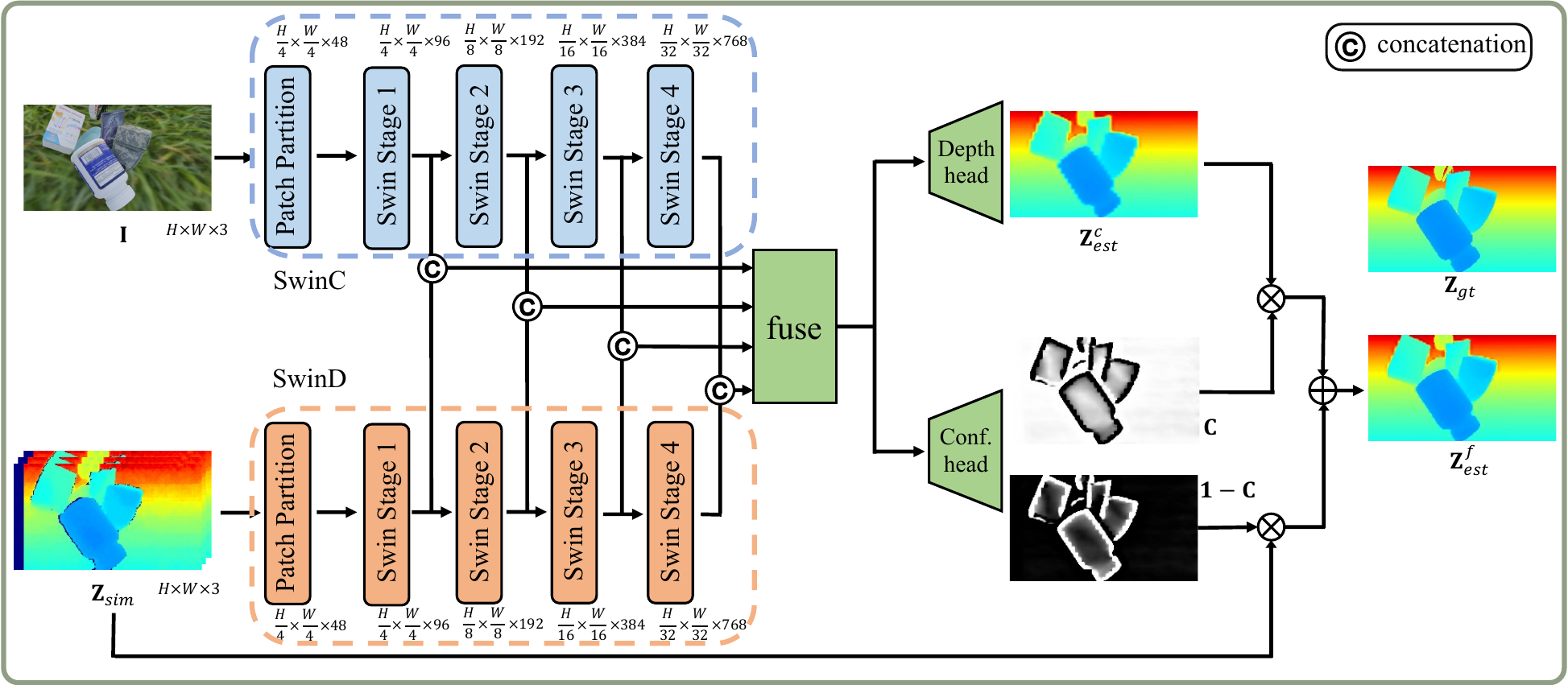}
\end{center}
\vspace{-3mm}
  \caption{The architecture of SDRNet.}
\label{fig:network}
\vspace{-3mm}
\end{figure*}

\section{DOWNSTREAM TASKS}

In this section, we introduce two downstream tasks, \ie~depth completion and depth pre-training, where our RaSim dataset effectively addresses the data scarcity issue and assists in real-world applications.

\subsection{Depth Completion}


As depicted in \fig{fig:network}, 
taking an RGB image $\mathbf{I} \in \mathbb{R}^{H \times W \times 3}$ and the repeated simulated depth $\mathbf{Z}_{sim} \in \mathbb{R}^{H \times W \times 3}$ as input, our SDRNet first respectively extracts hierarchical color and depth features with two Swin Transformer~\cite{liu2021swin} backbones named SwinC and SwinD.
The features are then concatenated and fed into two UPerNet~\cite{xiao2018unified} based decoders predicting a coarse depth map $\mathbf{Z}_{est}^c$ and a confidence map $\mathbf{C}$ as used in \cite{dai2022dreds}.
The final depth prediction is composed of the input and predicted depth as
\begin{equation}
\mathbf{Z}_{est}^f = ( 1 - \mathbf{C}) \otimes \mathbf{Z}_{sim} + \mathbf{C} \otimes \mathbf{Z}_{est}^c,
\end{equation}
where $\otimes$ denotes element-wise production.

Apart from the ground-truth depth, surface normal and gradient are also derived to supervise the coarse $\mathbf{Z}_{est}^c$ and fine $\mathbf{Z}_{est}^f$ depth estimation. 
The loss function is written as
\begin{equation}
\begin{aligned}
    \mathcal{L} &= \mathcal{L}^f + w_c \mathcal{L}^c,  \\
    \mathcal{L}^{\{c,f\}} &= \mathcal{L}_\mathbf{Z}^{\{c,f\}} + w_n \mathcal{L}_\mathbf{N}^{\{c,f\}} + w_g \mathcal{L}_\mathbf{G}^{\{c,f\}},
\end{aligned}
\end{equation}
where $\mathcal{L}_\mathbf{Z},\mathcal{L}_\mathbf{N},\mathcal{L}_\mathbf{G}$ denote the $\mathcal{L}_1$ losses of ground-truth and estimated depth, surface normal and gradient, and $w_c,w_n,w_g$ are the loss factors.
This optimization target enables $\mathbf{Z}_{est}^c$ to target the easily predictable area like the background while $\mathbf{Z}_{est}^f$ could focus on the challenging area like the edge of objects.

Notably, the network is trained with pure synthetic data and tested with the ClearGrasp dataset~\cite{sajjan2020clear} collected from the real world.

\subsection{Depth Pre-training}

\begin{figure}[tb!]
\begin{center}
\includegraphics[width=\linewidth]{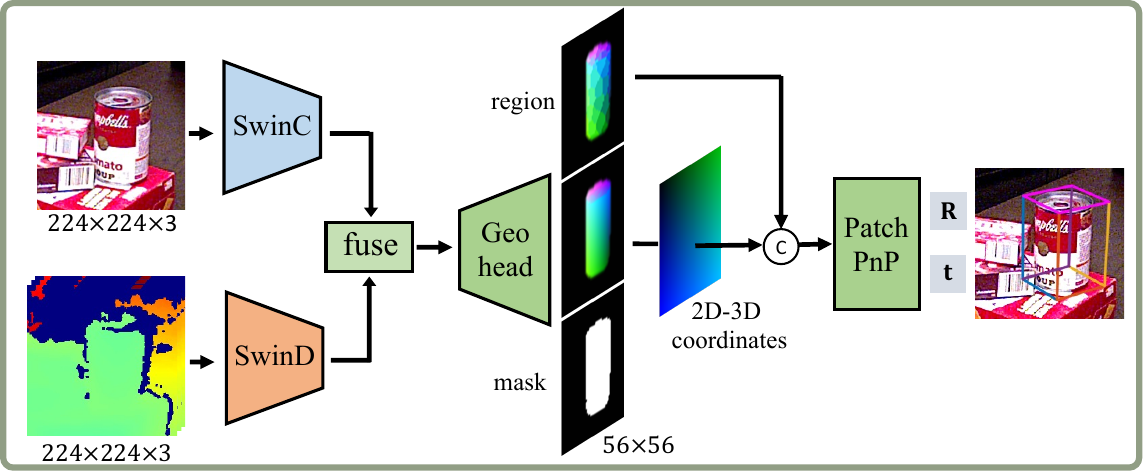}
\end{center}
\vspace{-3mm}
  \caption{The architecture for object pose estimation.}
\label{fig:pose}
\vspace{-3mm}
\end{figure}

To alleviate the lack of data, one de-facto paradigm is to pre-train neural networks on large-scale datasets and finetune on downstream task-specific datasets. 
Inspired by the idea of \textit{masked language modeling}, \ie~masking part of the data and then predicting the invisible content according to context, we introduce the simulated depth restoration as a pre-text task for depth-based Transformer pre-training.
Specifically, we first pre-train an SDRNet with the proposed RaSim dataset.
After pre-training, two homogeneous Swin backbones, \ie ~SwinC to encode color information and SwinD to encode depth information, are initialized by ImageNet-21K~\cite{imagnet21k} and our RaSim dataset pre-trained weights separately.
In this way, the pre-trained SwinD backbone gains prior knowledge of 3D geometric structures, thus benefiting various 3D tasks.

We choose object pose estimation as a verification task, for which collecting real-world annotations is oftentimes very expensive.
The objective of pose estimation is solving the 6DoF object pose, \ie ~3DoF rotation and 3DoF translation, in the camera coordinate system.
As shown in \fig{fig:pose}, the features extracted from zoomed-in RGB-D images are first aggregated and then sent to a geometric head to decode surface region, 3D coordinate map, and object mask. 
Afterwards, the surface region along with a 2D-3D dense correspondence map is fed into a Patch-PnP module to solve allocentric continuous rotation $\mathbf{R}$ and scale-invariant translation $\mathbf{t}$ as used in~\cite{Wang_2021_GDRN,di2021sopose,gao2022polarimetric}.

%% file: contents/5_exp.tex
\section{EXPERIMENTS}

\subsection{Depth Completion}
\noindent \textbf{Implementation Details.}
The model is trained on our RaSim dataset with the backbone of Swin-tiny (Swin-T).
To adapt the input scale of the backbone, we resize the depth map to $224\times224$ or $512\times512$ with nearest-neighbor interpolation.
We employ the Ranger optimizer~\cite{ranger1,ranger2,ranger3} with a learning rate of $1 \times 10^{-4}$ and a weight decay of 0.01.
The training epoch is set to 10 with a batch size of 32.

\noindent \textbf{Dataset.}
We evaluate SDRNet with the ClearGrasp~\cite{sajjan2020clear} test split, which contains 286 real-world RGB-D images of transparent objects along with their corresponding ground-truth depth maps.

\noindent \textbf{Evaluation Metrics.}
We follow the evaluation protocol of \cite{sajjan2020clear,zhu2021rgb}.
The predicted and ground-truth depth maps are first resized to $144\times256$, and we use four evaluation metrics: (1) root mean squared error (RMSE), (2) absolute relative difference (REL), (3) mean absolute error (MAE), and (4) the threshold $\delta$ which satisfies $\delta>\max(\frac{\tilde{d}_i}{d_i},\frac{d_i}{\tilde{d}_i})$, where $\delta \in \{1.05, 1.10, 1.25\}$, $d_i$ and $\tilde{d}_i$ denote ground-truth and predicted depths.

\input{tables/tab_cleargrasp}
\noindent \textbf{Comparison with State of the Arts.}
We compare our method with several top-performing methods in \tbl{tab:cleargrasp}.
Our SDRNet exceeds previous state-of-the-art methods~\cite{sajjan2020clear,zhu2021rgb} by a large margin and achieves comparable results with \cite{dai2022dreds}.
Note that \cite{sajjan2020clear,zhu2021rgb} are trained with data from ClearGrasp, while our network is trained exclusively on the synthetic RaSim dataset and demonstrates superior performance when transferred to real-world scenes.
These results confirm the high quality of the RaSim dataset and its effectiveness in bridging the sim-to-real domain gap.

\subsection{Depth Pre-training for Object Pose Estimation}

\noindent \textbf{Implementation Details.} 
The experiments are implemented with PyTorch~\cite{paszke2019pytorch}. 
Pose estimation models are trained for 12 epochs using the Ranger optimizer with a batch size of 24 and a learning rate of $1 \times 10^{-4}$, annealing at 50\% of the training phase leveraging a cosine schedule~\cite{loshchilov-ICLR17SGDR}.
The objects of interest are obtained using the detection results of YOLOX~\cite{ge2021yolox}.
In all experiments, one model is trained for all objects.

\noindent \textbf{Dataset.} 
The experiments are conducted on the widely used YCB-V~\cite{posecnn} dataset. 
It comprises more than 110K images in 92 RGB-D videos spanning 21 selected objects from the YCB object set. 
The dataset is challenging due to severe occlusions, symmetric objects, variable lighting conditions, and noisy depth.
Additionally, we also use the publicly available PBR images~\cite{hodan2020bop,hodan2019photorealistic} to aid training.

\noindent \textbf{Evaluation Metrics.}
We use the most common metric ADD and its variants for evaluation.
The error of ADD metric~\cite{Hinterstoisser2012Model,hodan2016evaluation} calculates the average distance of the object vertices transformed by the ground-truth pose [$\mathbf{R}|\mathbf{t}$] and the estimated pose [$\tilde{\mathbf{R}}|\tilde{\mathbf{t}}$]
\begin{equation}
\label{eq:add}
e_{\text{ADD}} = \frac{1}{N_{v}}\sum_{i=1}^{N_{v}} \|(\mathbf{R}\mathbf{x}_i+\mathbf{t})-(\tilde{\mathbf{R}}\mathbf{x}_i+\tilde{\mathbf{t}})\|.
\end{equation}
It is considered correct if $e_{\text{ADD}}$ is below 10\% of the object diameter.
For symmetric objects, the $e_{\text{ADD-S}}$ is employed based on the distance to the closest model point 
\begin{equation}
\label{eq:adds}
e_{\text{ADD-S}} = \frac{1}{N_{v}}\sum_{i=1}^{N_{v}} \mathop{\min}\limits_{\mathbf{x}_j \in \mathbb{V}} \|(\mathbf{R}x_i+\mathbf{t})-(\tilde{\mathbf{R}}\mathbf{x}_j+\tilde{\mathbf{t}})\|.
\end{equation}
We also report the area under curve (AUC) of ADD and ADD-S metrics by varying the distance threshold from 0cm to 10cm following \cite{posecnn}.

\begin{figure}[tb!]
\begin{center}
  \vspace{1mm}
  \includegraphics[width=0.85\linewidth]{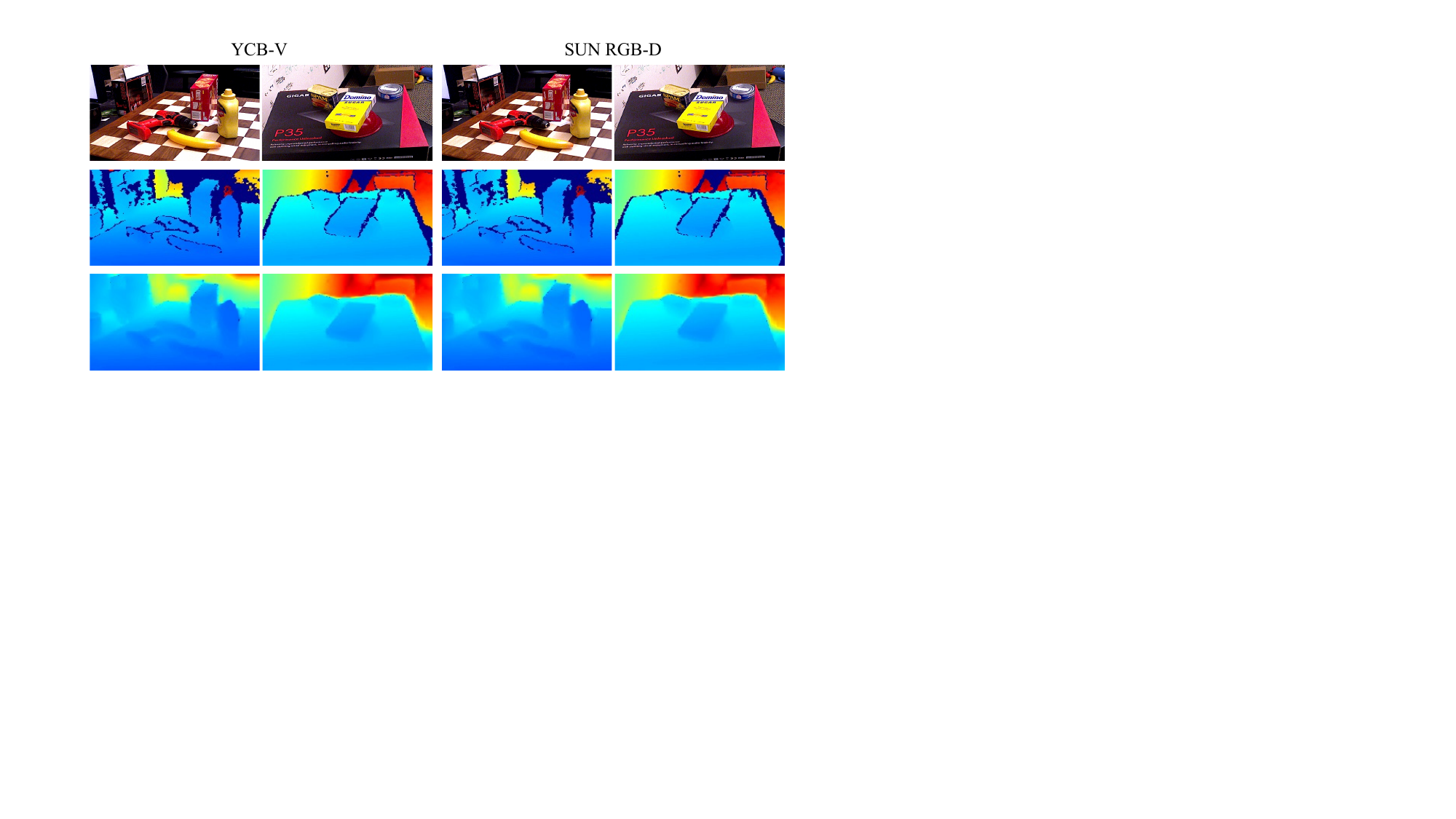}
\end{center}
\vspace{-3mm}
  \caption{Visualization results of depth restoration on YCB-V.}
\label{fig:depth_comp}
\vspace{-3mm}
\end{figure}

\noindent \textbf{Zero-shot Depth Restoration on YCB-V.}
To appraise whether the pre-trained model has a transfer ability on datasets in different domains, we perform depth restoration experiments on the YCB-V dataset.
Although the lack of ground-truth depth constraints the calculation of quantitative results, the qualitative results still reveal that the SDRNet trained from the synthetic dataset could generalize well in real-world scenarios, as shown in \fig{fig:depth_comp}.
It further proves the broad usage of the proposed RaSim dataset.

\input{tables/pose_sota}

\noindent \textbf{Comparison with State of the Arts.}
\tbl{tab:pose_sota} presents our quantitative results with other state-of-the-art methods ~\cite{wang2019densefusion,he2020pvn3d,he2021ffb6d,jiang2022uni6d} on YCB-V.
Our pre-trained model achieves 98.2\% on AUC of ADD-S and 93.8\% on AUC of ADD(-S), surpassing all the compared methods without any time-consuming refinement procedure.
Note that \cite{wang2019densefusion,he2021ffb6d} focus on designing complex fusion strategies for color and depth features, while \cite{jiang2022uni6d} directly concatenates the RGB and depth map and feeds them into the network.
However, our strategy is moderate yet more reasonable: leveraging the homogeneous Transformer backbones to extract multimodal features while initializing them with heterogeneous pre-trained weights.
This simplifies the network architecture yet maintains high accuracy and efficiency.

\input{tables/pose_abla}

\noindent \textbf{Ablation studies.}
\tbl{tab:pose_abla} illustrates several ablations \wrt depth pre-training strategies.
We can observe that initializing SwinD with ImageNet pre-trained weights brings slight enhancement over PyTorch's default random initialization (\tbl{tab:pose_abla} A1 \vs A0).
Nevertheless, our depth pre-training shows more distinct superiority, achieving an enhancement of 3.7\% on ADD(-S) metric and 2.9\% on the AUC of ADD(-S) metric (\tbl{tab:pose_abla} B0 \vs A0).

Aside from pre-training on RaSim, a simpler approach is applying randomization after rendering depth (Ren \& Ran), like adding Gaussian noise or randomly dropping depth values.
We observe this straightforward strategy is also effective (\tbl{tab:pose_abla} B1 \vs A0), but worse than the performance of RaSim initialization (\tbl{tab:pose_abla} B1 \vs B0).
This result reveals that the RaSim pipeline effectively shrinks the sim-to-real domain gap.

As mentioned in \sect{sec:dyren}, the range-aware rendering strategy broadens the depth range and enriches data diversity.
When the network is pre-trained solely on the stereo IR split (\tbl{tab:pose_abla} C0) or RGB split (\tbl{tab:pose_abla} C1), the performance drop distinctly on the ADD(-S) metric.


\begin{figure}[tb!]
\begin{center}
  \includegraphics[width=0.85\linewidth]{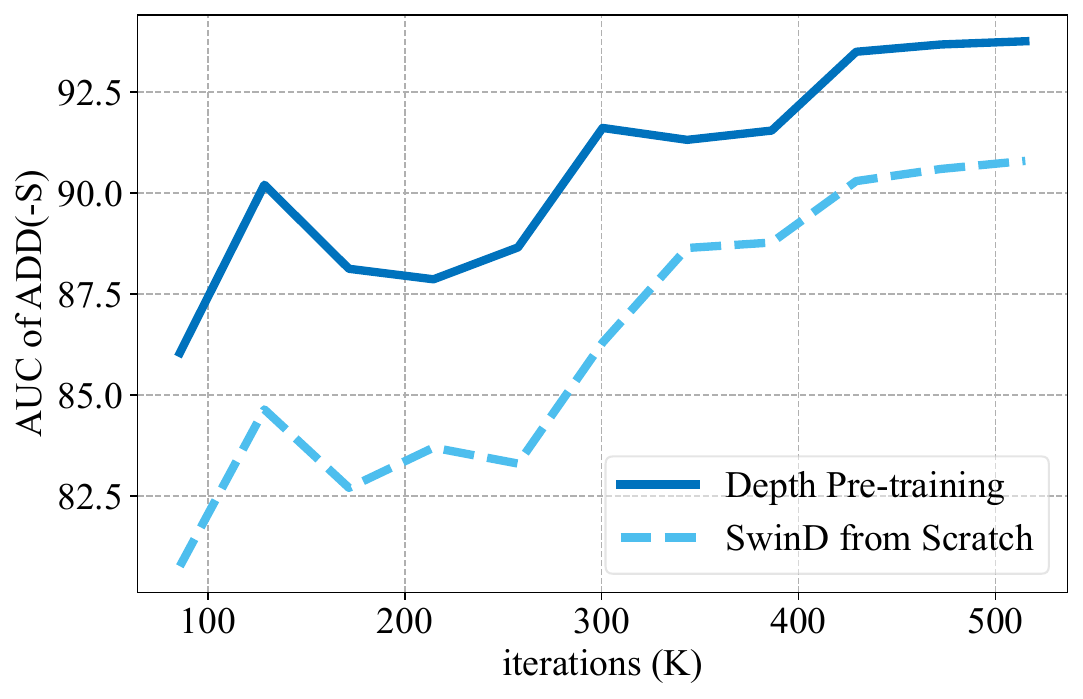}
\end{center}
  \caption{AUC of ADD(-S) \vs iterations on YCB-V.}
\vspace{-5mm}
\label{fig:add_iters}
\end{figure}

We illustrate the results of the baseline and our pre-training \vs iterations in \fig{fig:add_iters}.
As is depicted, pre-training with the RaSim significantly boosts the performance, especially in the early stage of training.
This indicates that pre-training on the RaSim dataset effectively equips the Transformer-based backbone with prior 3D geometric knowledge.

%% file: tables/tab_cleargrasp.tex
\begin{table}[tbp]
\vspace{1mm}
\caption{\textbf{ Comparison with state-of-the-art methods on ClearGrasp.} $\downarrow$ means lower is better, $\uparrow$ means higher is better. RGBD-FCN and \cite{park2020non} are implemented by~\cite{zhu2021rgb}.}
\setlength\tabcolsep{3.2pt}
\centering
\begin{tabular}{l|cccccc}
\shline
\multicolumn{1}{c|}{Methods} & RMSE$\downarrow$ & REL$\downarrow$ & MAE$\downarrow$  & $\delta_{1.05}\uparrow$ & $\delta_{1.10}\uparrow$ & $\delta_{1.25}\uparrow$ \\
\hline
\multicolumn{1}{c|}{} & \multicolumn{6}{c}{ClearGrasp Real-known} \\
\hline
RGBD-FCN & 0.054 & 0.087 & 0.048 & 36.32 & 67.11 & 96.26 \\
NLSPN~\cite{park2020non}     & 0.149 & 0.228 & 0.127 & 14.04 & 26.67 & 54.32 \\
CG~\cite{sajjan2020clear}    & 0.039 & 0.051 & 0.029 & 72.62 & 86.96 & 95.58 \\
LIF~\cite{zhu2021rgb}        & 0.028 & 0.033 & 0.020 & 82.37 & 92.98 & 98.63 \\
DREDS~\cite{dai2022dreds}    & 0.022 & \textbf{0.017} & 0.012 & 91.46 & \textbf{97.47} & \textbf{99.86}\\
Ours                         & \textbf{0.021} & \textbf{0.017} & \textbf{0.011} & \textbf{94.14} & \textbf{97.47} & 99.58 \\
\hline
\multicolumn{1}{c|}{} & \multicolumn{6}{c}{ClearGrasp Real-novel} \\
\hline
RGBD-FCN & 0.042 & 0.070 & 0.037 & 42.45 & 75.68 & 99.02 \\
NLSPN~\cite{park2020non}  & 0.145 & 0.240 & 0.123 & 13.77 & 25.81 & 51.59 \\
CG~\cite{sajjan2020clear} & 0.034 & 0.045 & 0.025 & 76.72 & 91.00 & 97.63 \\
LIF~\cite{zhu2021rgb}     & 0.025 & 0.036 & 0.020 & 76.21 & 94.01 & 99.35 \\
DREDS~\cite{dai2022dreds} & 0.016 & \textbf{0.008}  & \textbf{0.005} & \textbf{96.73} & \textbf{98.83} & 99.78 \\ 
Ours                      & \textbf{0.014} & 0.010 & \textbf{0.005} & 95.74 & 98.26 & \textbf{99.87} \\
\shline
\end{tabular}
\label{tab:cleargrasp}
\vspace{-3mm}
\end{table}

%% file: tables/pose_sota.tex
\begin{table*}[tbp]
\vspace{1mm}
\caption{
\label{tab:pose_sota}
{ \bf Comparison with state of the arts on YCB-V.}
Here ADD(-S) uses the symmetric metric only for symmetric objects (denoted with $^*$), while ADD-S uses the symmetric metric for all objects.
}
\centering
\tablestyle{4pt}{1.1}
\resizebox{0.95\linewidth}{!}{
\begin{tabular}{l|cc|cc|cc|cc|cc}
    \shline
  & \multicolumn{2}{c|}{DenseFusion~\cite{wang2019densefusion}} & \multicolumn{2}{c|}{PVN3D~\cite{he2020pvn3d}}  & \multicolumn{2}{c|}{FFB6D~\cite{he2021ffb6d}} & \multicolumn{2}{c|}{Uni6D~\cite{jiang2022uni6d}}  & \multicolumn{2}{c}{Ours} \\ 
 \hline 
 Object & \tabincell{c}{AUC of\\ADD-S} & \tabincell{c}{AUC of\\ADD(-S)} & \tabincell{c}{AUC of\\ADD-S} & \tabincell{c}{AUC of\\ADD(-S)} & \tabincell{c}{AUC of\\ADD-S} & \small{\tabincell{c}{AUC of\\ADD(-S)}} & \tabincell{c}{AUC of\\ADD-S} & \tabincell{c}{AUC of\\ADD(-S)} &\tabincell{c}{AUC of\\ADD-S} & \tabincell{c}{AUC of\\ADD(-S)} \\
 \hline 
        002\_master\_chef\_can  & 95.3 & 70.7 & 96.0 & 80.5 & 96.3 & \textbf{80.6} & 95.4 & 70.2 & \textbf{100.0} & 80.2\\
        003\_cracker\_box       & 92.5 & 86.9 & 96.1 & \textbf{94.8} & \textbf{96.3} & 94.6 & 91.8 & 85.2 & 92.1 & 71.9\\ 
        004\_sugar\_box         & 95.1 & 90.8 & 97.4 & 96.3 & 97.6 & 96.6 & 96.4 & 94.5 & \textbf{99.9} & \textbf{99.6}\\ 
        005\_tomato\_soup\_can  & 93.8 & 84.7 & 96.2 & 88.5 & 95.6 & 89.6 & 95.8 & 85.4 & \textbf{98.7} & \textbf{96.7}\\
        006\_mustard\_bottle    & 95.8 & 90.9 & 97.5 & 96.2 & 97.8 & 97.0 & 95.4 & 91.7 & \textbf{100.0} & \textbf{100.0}\\   
        007\_tuna\_fish\_can    & 95.7 & 79.6 & 96.0 & 89.3 & 96.8 & 88.9 & 95.2 & 79.0 & \textbf{100.0} & \textbf{98.8}\\ 
        008\_pudding\_box       & 94.3 & 89.3 & 97.1 & \textbf{95.7} & 97.1 & 94.6 & 94.1 & 89.8 & \textbf{99.2} & 92.7\\ 
        009\_gelatin\_box       & 97.2 & 95.8 & 97.7 & 96.1 & 98.1 & 96.9 & 97.4 & 96.2 & \textbf{100.0} & \textbf{99.9}\\  
        010\_potted\_meat\_can  & 89.3 & 79.6 & 93.3 & 88.6 & 94.7 & 88.1 & 93.0 & 89.6 & \textbf{95.9} & \textbf{90.6}\\ 
        011\_banana             & 90.0 & 76.7 & 96.6 & 93.7 & 97.2 & 94.9 & 96.4 & 93.0 & \textbf{100.0} & \textbf{99.2}\\ 
        019\_pitcher\_base      & 93.6 & 87.1 & 97.4 & 96.5 & 97.6 & 96.9 & 96.2 & 94.2 & \textbf{100.0} & \textbf{99.8}\\  
        021\_bleach\_cleanser   & 94.4 & 87.5 & 96.0 & 93.2 & 96.8 & 94.8 & 95.2 & 91.1 & \textbf{99.1} & \textbf{95.8}\\  
        024\_bowl$^*$           & 86.0 & 86.0 & 90.2 & 90.2 & \textbf{96.3} & \textbf{96.3} & 95.5 & 95.5 & 94.0 & 94.0\\ 
        025\_mug                & 95.3 & 83.8 & \textbf{97.6} & \textbf{95.4} & 97.3 & 94.2 & 96.6 & 93.0 & 96.1 & 95.0\\  
        035\_power\_drill       & 92.1 & 83.7 & 96.7 & 95.1 & 97.2 & 95.9 & 94.7 & 91.1 & \textbf{99.9} & \textbf{96.4}\\  
        036\_wood\_block$^*$    & 89.5 & 89.5 & 90.4 & 90.4 & 92.6 & 92.6 & 94.3 & 94.3 & \textbf{96.9} & \textbf{96.9}\\ 
        037\_scissors           & 90.1 & 77.4 & 96.7 & 92.7 & \textbf{97.7} & \textbf{95.7} & 87.6 & 79.6 & 94.6 & 70.6\\  
        040\_large\_marker      & 95.1 & 89.1 & 96.7 & 91.8 & 96.6 & 89.1 & 96.7 & 92.8 & \textbf{99.8} & \textbf{94.3}\\
        051\_large\_clamp$^*$   & 71.5 & 71.5 & 93.6 & 93.6 & 96.8 & 96.8 & 95.9 & 95.9 & \textbf{99.2} & \textbf{99.2}\\  
 052\_extra\_large\_clamp$^*$   & 70.2 & 70.2 & 88.4 & 88.4 & 96.0 & 96.0 & 95.8 & 95.8 & \textbf{97.6} & \textbf{97.6}\\  
         061\_foam\_brick$^*$   & 92.2 & 92.2 & 96.8 & 96.8 & 97.3 & 97.3 & 96.1 & 96.1 & \textbf{99.9} & \textbf{99.9}\\ 
        \hline
        Avg                     & 91.2 & 82.9 & 95.5 & 91.8 & 96.6 & 92.7 & 95.2 & 88.8 & \textbf{98.2} & \textbf{93.8}\\
        \shline
\end{tabular}}
\vspace{-4mm}
\end{table*}

%% file: tables/pose_abla.tex
\begin{table}
\caption{
\label{tab:pose_abla}
{\bf Ablations on depth pre-training strategies.}
We report the results of ADD(-S), and AUC of ADD-S and ADD(-S) metrics on the YCB-V dataset.
Ren \& Ran is short for randomization after rendering depth.
}
\centering
\scalebox{0.94}{
\tablestyle{5pt}{1.2}
\begin{tabular}{@{}l|cc|ccc@{}}
    \shline
    Row & SwinC & SwinD & ADD(-S)  & \tabincell{c}{AUC of\\ADD(-S)} & \tabincell{c}{AUC of\\ADD-S}  \\ 
    \hline
    A0  & ImageNet-21K & Random           & 81.8 & 90.9 & 97.6 \\
    A1  & ImageNet-21K & ImageNet-21K         & 82.5 & 92.8 & 97.8\\ 
    \hline
    B0  & ImageNet-21K & RaSim      & \textbf{85.5} & \textbf{93.8} & \textbf{98.2}\\
    B1  & ImageNet-21K & Ren \& Ran       & 84.6 & 93.3 & 97.6\\
    \hline
    C0  & ImageNet-21K & Stereo IR Split     & 83.9 & 93.4 & 98.0 \\
    C1  & ImageNet-21K & Stereo RGB Split    & 83.7 & 93.6 & 97.9 \\
    \shline
\end{tabular}
}
\vspace{-3mm}
\end{table}

%% file: contents/6_conclusion.tex
\section{CONCLUSION}

This work has introduced RaSim, a range-aware RGB-D data simulation pipeline that excels in producing high-fidelity RGB-D data.
By imitating the imaging principle of real-world depth sensors, we effectively bridge the sim-to-real domain gap concerning depth maps.
Notably, we incorporate a range-aware rendering strategy to enrich data diversity, making RaSim generalizable to a broader range of real-world application scenarios.
Experiments on 3D perception tasks demonstrate that models trained with RaSim can be directly applied to real-world datasets like ClearGrasp and YCB-V without the need for finetuning.
In the future, we aim to explore the simulation of more types of depth sensors and expand RaSim to more diverse applications.

%% file: contents/7_acknowledgment.tex

\noindent \textbf{Acknowledgments.}
This work was supported by the National Key R\&D Program of China under Grant 2018AAA0102801.